\documentclass[conference]{IEEEtran}
\IEEEoverridecommandlockouts
\usepackage{flushend}
\usepackage{cite}
\usepackage{amsmath,amssymb,amsfonts}
\usepackage{algorithmic}
\usepackage{graphicx}
\usepackage{textcomp}
\usepackage{orcidlink}
\usepackage[acronym,toc]{glossaries}
\newacronym{snn}{SNN}{Spiking Neural Network}
\newacronym{lsm}{LSM}{Liquid State Machine}
\newacronym{iaf}{IaF}{Integrate-and-Fire}
\newacronym{lif}{LIF}{Leaky-Integrate-and-Fire}
\newacronym{stdp}{STDP}{Synaptic Time Dependent Plasticity}
\newacronym{vp}{VP}{ Victor Purpura}
\newacronym{vr}{VR}{Van Rossum}
\newacronym{ann}{ANN}{Artificial Neural Network}
\newacronym{sap}{SAP}{Spike After Potential}
\newacronym{ipsp}{IPSP}{Inhibitory Post Synaptic Potential}
\newacronym{epsp}{EPSP}{Excitatory Post Synaptic Potential}
\newacronym{ai}{AI}{Artificial Intelligence}
\makeglossaries
\usepackage{xcolor}
\def\BibTeX{{\rm B\kern-.05em{\sc i\kern-.025em b}\kern-.08em
 T\kern-.1667em\lower.7ex\hbox{E}\kern-.125emX}}
\begin{document}

\title{Weights Initialisation of Liquid State Machines}

\author{
\IEEEauthorblockN{Pavithra Koralalage\IEEEauthorrefmark{1}, Ireoluwa Fakeye\IEEEauthorrefmark{1}, Pedro Machado\IEEEauthorrefmark{1}\orcidlink{0000-0003-1760-3871}, Jason Smith\IEEEauthorrefmark{2}\orcidlink{0000-0002-4209-1604}, Isibor Kennedy Ihianle\IEEEauthorrefmark{1}\orcidlink{0000-0001-7445-8573},\\ Salisu Wada Yahaya\IEEEauthorrefmark{1} and Andreas Oikonomou\IEEEauthorrefmark{1}\orcidlink{0000-0002-0394-6112}}
\IEEEauthorblockA{\IEEEauthorrefmark{1} Department of Computer Science, Nottingham Trent University\\
Nottingham, NG11 8NS, UK\\
Emails: \{N1043138,N1112313\}@my.ntu.ac.uk,\\\{pedro.machado, isibor.ihianle, salisu.yahaya, andreas.oikonomou\}@ntu.ac.uk}
\IEEEauthorblockA{\IEEEauthorrefmark{2}Department of Maths, Nottingham Trent University\\
Nottingham, NG11 8NS, UK\\
Email: jason.smith@ntu.ac.uk}
}

\maketitle

\begin{abstract}
\glspl*{snn} emerged as a promising solution in the field of \glspl*{ann}, attracting the attention of researchers due to their ability to mimic the human brain and process complex information with remarkable speed and accuracy. This research aimed to optimise the training process of \glspl*{lsm}, a recurrent architecture of \glspl*{snn}, by identifying the most effective weight range to be assigned in \gls*{snn} to achieve the least difference between desired and actual output. The experimental results showed that by using spike metrics and a range of weights, the desired output and the actual output of spiking neurons could be effectively optimised, leading to improved performance of \glspl*{snn}. The results were tested and confirmed using three different weight initialisation approaches, with the best results obtained using the Barabasi-Albert random graph method.

\end{abstract}

\begin{IEEEkeywords}
Spiking Neural Networks (\gls*{snn}), Liquid State Machine (\gls*{lsm}), Victor Purpura Distance, Van Rossum Distance, Excitatory Synapses, Inhibitory Synapses, NEST, Weight initialisation, random weights, Barabasi-Albert graph, Erdos Renyi graph
\end{IEEEkeywords}

\section{Introduction}
The evolving journey of \acrfull*{ann} is based on attempting to mimic the behaviour of biological neurons. 
The capability of the mammalian brain to understand and process remarkable information within an unbelievably short amount of time is taken into consideration when enhancing the neural networks.\acrfull*{snn} has evolved based on the behaviour of the mammalian visual cortex. The visual cortex is mainly responsible for advanced image processing within a short period of time. Similar to the behaviour of the visual cortex of the mammalian brain, \gls*{snn} use spikes over time as the input for the network, and neurons spike when they reach their threshold value. This mimicked behaviour capable of offering low-power and high-performance computational paradigm \cite{machado2019, schliebs2013, auge2021}.

Liquid State Machine (\glspl*{lsm}) are a type of recurrent \gls*{snn} that is considered to be more biologically accurate than other neural network methods \cite{fountain2022}. \glspl*{lsm} are well-known for their computational power and low energy consumption, making them a commonly used tool for complex classification tasks such as speech recognition and image classification. Optimising the training process is a crucial aspect in the development of \glspl*{snn}, but it still lacks proper research attention. The effective assignment of weight ranges in \glspl*{snn} plays a crucial role in achieving accurate solutions that are also energy efficient. 

This article presents a comprehensive investigation into optimising the training process of the \gls*{lsm}. The study sheds light on how to make use of spike metrics to compare spike trains and determine the accuracy of the output. The article also highlights the importance of balanced excitatory and inhibitory synapses in achieving optimal output. In addition, the study explains the significance of weight initialisation, which involves assigning initial weights to the synapses, and how it can impact the performance of the \gls*{lsm}. Through experiments with different weight ranges, the article identifies the most effective weight range for the \gls*{lsm}, making it a valuable resource for researchers and practitioners looking to optimise their training processes.

The article is organised as follows: Section \ref{sec:lr} presents a thorough review of the relevant literature in the field. The methodology used in the study is described in detail in Section \ref{sec:method}. The results of the analysis are presented in Section \ref{sec:results}, where the findings are discussed and interpreted. Finally, the article concludes in Section \ref{sec:conclusions} which summarises the main contributions of the study and outlines potential future work in the field.

\section{Literature Review} \label{sec:lr}
\glspl*{snn} have emerged as the third generation of \glspl*{ann}, surpassing the limitations of traditional neural networks 
\cite{zhang2021}. \glspl*{snn} mimic the biological processes of the human brain, making them capable of providing high-performance solutions with low energy consumption. These networks have proven to solve complex problems that were difficult to address with classical neural networks \cite{auge2021, gavrilov2016}.

Traditional neural networks rely on stochastic gradient descent algorithms, which require all neurons' weights to be adjusted in each iteration through the backpropagation algorithm. This constant adjustment of neurons' weights requires a significant amount of power and time. On the other hand, \glspl*{snn} only activate neurons that reach their threshold values. The communication between neurons in \glspl*{snn} is also done via discrete spikes over time, rather than the continuous flow of information in traditional neural networks. This activation process requires much less energy and time, making \glspl*{snn} an efficient alternative to traditional neural networks \cite{hwang2021}.

\begin{figure}[!t]
\centering
\includegraphics[width=0.5\textwidth]{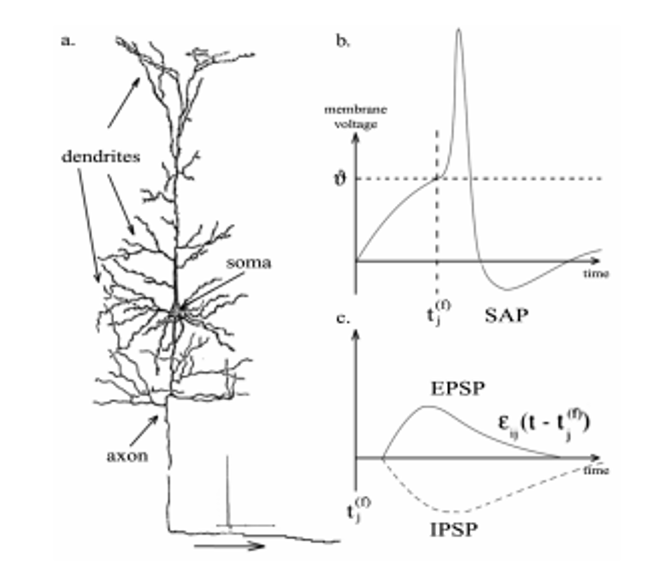}
\caption{a. biological neuron, b. incoming \acrfull*{sap}, c. behaviour of \acrfull*{ipsp} and \acrfull*{epsp} \cite{vreeken2003}. \label{fig:biological}}
\end{figure}

An illustration of biological neurons with incoming post-synaptic potentials including incoming \gls*{sap} and  behaviour of \gls*{ipsp} and \gls*{epsp} are shown in Fig. \ref{fig:biological}.

A key characteristic of \gls*{snn} is their ability to mimic the behaviour of the mammalian brain and process complex information quickly and accurately. This is achieved through their utilisation of various network architectures, including feed-forward, recurrent, and hybrid models \cite{zhang2021}. Among these, the feed-forward architecture is the most commonly used due to its simplicity and ease of transitioning from traditional neural networks \cite{gupta2020}. A basic \gls*{snn} structure may consist of three layers, with the hidden layer comprised of spiking neurons. Input information can be converted into spike trains through encoding techniques such as rate coding and temporal coding \cite{mohemmed2013,hwang2021}.

The \gls*{iaf} is a simplistic spiking neuron model that treats the neuron as an electronic circuit. The membrane potential is calculated based on the synaptic input and the input current, and spikes are generated when the membrane potential reaches the threshold. This model is similar to the \gls*{lif} model, but with a focus on the neuron's ability to mimic the spike-generating behaviour of real neurons, rather than the leaky behaviour of traditional electronic circuits \cite{burkitt2006}.

\begin{figure}[!t]
\centering
\includegraphics[width=0.5\textwidth]{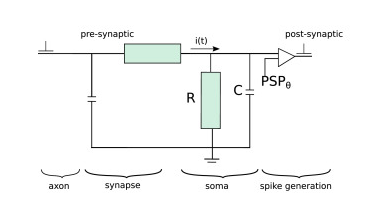}
\caption{Leaky-Integrate-and-Fire model \cite{lobo2020}.}\label{fig:lif}
\end{figure}

The \gls*{lif} model shown in Fig. \ref{fig:lif}, on the other hand, is a widely-used spiking neuron model that is popular due to its low computational cost, ease of coding, and high degree of accuracy. The \gls*{lif} model captures the characteristic properties of external input based on the charge across a leaky membrane with a clear threshold, making it a one-dimensional \gls*{snn}. Like the \gls*{iaf} model, the \gls*{lif} neuron is considered as an electronic circuit and simulates the potential of the neuron as a voltage across a capacitor connected in parallel in a leaky conductance path. The dynamics of the membrane potential u(t) of a single neuron can be calculated using the following equation:

\begin{equation}
 \frac{du(t)}{dt}=\frac{u_{rest} - u(t)}{\tau m}+\frac{I(t)}{c_m} 
\end{equation}

where $\tau_{m}$ is membrane time constant, $ u_{rest} $ is resting potential $c_m $ is membrane capacitance and $I(t)$ is neuron input 
\cite{taherkhani2020,tavanaei2019}

Realistic neuron models such as the Hodgkin-Huxley and FitzHugh-Nagumo \cite{lobo2020} offer greater biological accuracy, but come at the cost of increased computational demands. On the other hand, the Izhikevich model \cite{izhikevich2003} strikes a balance between biological realism and computational efficiency, offering more biological accuracy than the Hodgkin-Huxley model while still having a computational cost similar to the simpler \gls*{lif} model. However, these complex models may not be suitable for future applications that require acceleration of the \gls*{snn} using dedicated parallelisable hardware, due to their higher computational demands.

\begin{figure}[!t]
\centering
\includegraphics[width=0.9\columnwidth]{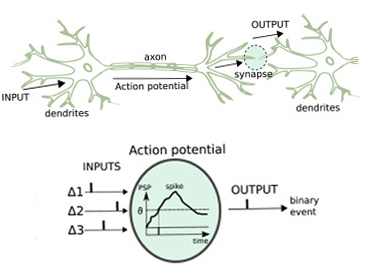}
\caption{ Biological neuron and association with artificial spiking neuron \cite{lobo2020}}\label{fig:syn}
\end{figure}

The behaviour of a spiking neuron is determined by the type of connections it has with other neurons, which can be classified as either excitatory or inhibitory. Excitatory connections positively impact the network's output, and the connected neurons are known as excitatory neurons. In contrast, inhibitory connections negatively impact the network's output and the connected neurons are referred to as inhibitory neurons. Fast spiking neurons emit consistent, high-frequency spikes, while Low-threshold spikes emit high-frequency spikes with varying frequency. The behaviour of spiking neurons can be depicted in Fig. \ref{fig:syn} with compared to biological neuron. 

The crucial role of excitatory and inhibitory neurons in the functioning of \glspl*{snn} lies in their ability to identify relevant features and activate only the necessary neurons with the appropriate threshold, thereby increasing the efficiency of the \glspl*{snn} \cite{izhikevich2003}.

The method of weight initialisation for synapses is a crucial factor in the performance of a \gls*{snn}. The weight values of synapses are among the most critical parameters that shape the behaviour of neurons within the network. There are various techniques for initialising the weight values of synapses, including random weight assignment within a specified range, the use of random graph models, etc. One such model is the Barabasi-Albert Graph, proposed by A. Barabasi and R. Albert in 1999. This model depicts the random growth of a network by adding edges to existing vertices in a manner proportional to the number of connections of a given vertex. Essentially, the more connections a vertex has, the higher the likelihood of it receiving new connections \cite{zadorozhnyi2012}. On the other hand, the Erdos Renyi Graph is one of the earliest forms of random graph models. This graph is created by randomly connecting nodes with a specified probability of each node \cite{gomez2006}. Unlike the Barabasi-Albert Graph, the Erdos Renyi Graph does not exhibit preferential connections between nodes. In this article, three main weight initialisation techniques were employed. The first approach involved randomly assigning weights within a range for the synapses within the \gls*{snn}. The Barabasi-Albert graph and the Erdos Renyi random graph were then utilised as the next methods for assigning weights to the synapses.

The role of Lateral Inhibition in optimising the performance of \gls*{snn} is a crucial topic of discussion. Lateral Inhibition is a biologically-inspired mechanism in the retina, which is responsible for edge detection and wide-range light intensity processing, as explained by Bowes et al. \cite{bowes2009}. Electrophysiological experiments by Hartline and his research team identified the lateral inhibition process in the neural circuit \cite{fang2007}. These experiments revealed that receptors in the network are inhibited by their adjacent receptors, and inhibition among cells is mutual. The inhibition from nearby receptors is stronger compared to that from far away receptors, leading to inconsistent receptor activities.

In \glspl*{snn}, lateral inhibition works by sending signals to other neurons to prevent them from firing for a targeted spike, thus allowing only the relevant neurons to be activated and reducing information overload \cite{hwang2021}. This mechanism reduces the need for membrane reset functions and helps to prevent neurons from over-saturation. Zhao et al. \cite{zhao2020} discussed a method to reduce the complexity of the lateral inhibition connection in the \gls*{snn}, leading to increased scalability compared to the two commonly used methods, which are challenging to implement. These two methods are: (1) every excitatory neuron is connected to a relevant inhibitory neuron and (2) every inhibitory neuron is connected to all excitatory neurons, except the neuron receiving the spike. Zhang et al. \cite{zhang2021} proposed a conceptual diagram for the use of lateral inhibition in the \gls*{snn} architecture, using backpropagation to calculate errors and adjust weights. As described by Zhang et al. \cite{zhang2021}, lateral inhibition in the \gls*{snn} reduces the activities of all other neurons once a desired output spike is generated. 

Learning algorithms are also crucial in projecting inputs towards a desired output and for accurately identifying important features. Unlike classical neural networks, the learning algorithms used in \gls*{snn} are based on spike trains and the time between each spike, rather than stochastic gradient descent. Three commonly used learning algorithms in \gls*{snn} are Hebbian learning, Oja learning, and \gls*{stdp}. Hebbian learning is a well-known and established method for training neural networks, and it is unique in that it is not based on gradient descent. Instead, weights are based on the correlation between inputs and outputs. This makes Hebbian learning more biologically plausible and more efficient for situations where gradient descent is not possible, and it can be used for both supervised and unsupervised tasks. Oja learning builds upon the Hebbian approach but addresses the issue of unbound weight growth, which can contradict experimental results in neural networks. Oja learning suggests normalising the synaptic weights before using them in the network to provide a fair range of values for learning. \gls*{stdp} is a Hebbian-based unsupervised learning algorithm that offers several advantages over classical Neural Networks learning algorithms. It is based on local event-based data and does not require additional energy or a large set of annotated data. In \gls*{stdp}, synapse weights are determined by the time difference between pre-spikes and post-spikes of neurons. It is mainly used to train excitatory neurons and classify them as critical or non-critical neurons. The weight update can be calculated using the equation provided in the original text.

\begin{equation}
\Delta w_j=\sum_{f=1}^{N}\sum_{n=1}^{N} W(t_{n}^{i} - t_{f}^{j})
\end{equation}

The alteration in the weight $\Delta w_{j}$ of a synapse connecting presynaptic neuron $j$ is based on the correlation between the timing of presynaptic spike arrivals and postsynaptic spikes. The arrival times of presynaptic spikes at synapse $j$ are represented as $ t_{f}^{j} $, where $f=1,2,3,...$ denotes the number of presynaptic spikes. Similarly, the firing times of the postsynaptic neuron are represented as $ t_{n}^{i} $, where $n=1,2,3,...$ labels the firing events. The overall weight change $\Delta w_{j}$ that results from a stimulation protocol that involves a sequence of pre and postsynaptic spikes is expressed as \cite{gerstner1996}.

\begin{equation}
W(x)=A^{+}+ exp(-x/\tau^{+}) , for x > 0
\end{equation}

The \gls*{stdp} function $W(x)$ determines the degree of synaptic strengthening or weakening based on the relative timing between presynaptic and postsynaptic spikes. It is a crucial component of the \gls*{stdp} learning algorithm, and its selection can greatly impact the performance of the network. A widely used form of the \gls*{stdp} function is the exponential function, which has been shown to effectively model the synaptic plasticity observed in biological neurons \cite{gerstner1996}.

\begin{equation}
W(x)=-A^{-}+ exp(-x/\tau^{-}) , for x < 0
\end{equation}

The parameters $A^{+}$ and $A^{-}$ may vary depending on the current weight $w_{j}$ of the synapse. The time constants of the \gls*{stdp} functions are typically around $10ms$ for both $\tau^{+}$ and $\tau^{-}$ \cite{gerstner1996}.

The alteration in weight is determined by the timing relationship between the arrival of the presynaptic spike and the postsynaptic spike. If the postsynaptic spike occurs directly after the presynaptic spike, the weight will increase. On the other hand, if there is a considerable time gap between the two spikes, the weight will decrease. The weight updates are based on the importance of the neurons, determined by their weight increase. If a neuron has a noticeable weight increase, it is considered critical and its weight is updated. Conversely, if a neuron does not show a significant weight increase, it is considered non-critical and its weight is set to zero. During the training process, the training data set is divided into multiple batches. The weights of critical neurons are updated, while the weights of non-critical neurons are set to zero. However, these non-critical neurons can still contribute to later batches, as they may become critical for different input sets. This approach makes the network scalable and stable, as reported in various studies \cite{rathi2018, wang2016, gavrilov2016}.

The properties and patterns of spikes and spike trains play a crucial role in determining the similarities and differences among spikes. This is essential for capturing important data within a spiking neural network to achieve optimised results. Victor et al. \cite{victor2005} categorise spike metrics into three types: spike-time metrics, spike-interval metrics, and multi-neuronal cost-based metrics. The choice of metric depends on the cost of inserting and deleting inter-spike intervals and the need to identify spike patterns as a series of events. Dauwels et al. \cite{dauwels2008} and Kreuz et al. \cite{kreuz2007} discuss two main spike train metrics. The Victor-Purpura Spike Train Metric calculates the distance between two point processes through minimum cost transformations using event insertion, deletion, and movement operations. The cost of event insertion or deletion is set to one and movement is proportional to the time change \cite{victor1996nature}. The Van Rossum Similarity Measure converts the two point processes into continuous time series. The Schreiber et al. \cite {schreiber2003new} Similarity Measure is based on correlation and involves convolving each spike with a filter of fixed width. The Hunter-Milton Similarity Measure begins from the nearest spike and is calculated as the average of the entire series \cite{hunter2003amplitude}.

\glspl*{lsm} can be defined as a biologically inspired recurrent spiking neural network that has high computational power and is widely used for tasks such as speech recognition, image classification, and word recognition. When compared to classical neural networks, \glspl*{lsm} performs well with high accuracy and low power consumption. \glspl*{lsm} are composed of three main components: the input layer, the liquid layer, and the read-out layer \cite{das2021,fountain2022,kang2022,tian2021,wang2016}. The input layer converts the input spike trains into spikes using encoding mechanisms, such as rate coding, time coding, and phase coding, and the weights are initialised using a fixed weight matrix. The liquid layer, consisting of \gls*{lif} neurons, extracts random features from the inputs, and there are two main types of neurons within the liquid layer: primary and auxiliary neurons, which are divided based on their role in the feature extraction process. The read-out layer provides the final output of the network and is responsible for training the liquid layer to read-out layer connected neurons weights. To optimise the training process of \glspl*{lsm}, a supervised-learning algorithm named ReSuMe can be used \cite{das2021,fountain2022,kang2022,tian2021,wang2016}. The ReSuMe algorithm trains the weights in the read-out layer based on the desired output and the actual output of the network \cite{ponulak2010}. During the training phase, the ReSuMe algorithm adjusts the weights in the read-out layer to minimise the difference between the desired output and the actual output. The algorithm performs this by updating the weights based on the error gradient, calculated using the backpropagation algorithm, and the learning rate, which determines the speed of the weight update. The ReSuMe algorithm is an effective solution for optimising the training process of \gls*{lsm} and improving its accuracy and performance.

Finally, the training process of \gls*{lsm} can be performed using the supervised-learning ReSuMe algorithm or unsupervised learning \gls*{stdp}. When the expected patterns are known, ReSuMe can be used to train the network by updating the weights from the liquid layer to the read-out layer. ReSuMe uses a suitable learning algorithm to adjust the weights and provide the desired output. On the other hand, when the expected patterns are not known, unsupervised learning \gls*{stdp} can be used to train the network by using \gls*{stdp} rules to adjust the weights. In conclusion, \gls*{lsm} has proven to be a powerful and efficient tool for processing complex information with high accuracy and low energy consumption.

\section{Methodology}\label{sec:method}
As part of the methodology for this study and experiment setting, the selection of a \gls*{snn} simulator is a crucial aspect. NEST \cite{NEST} was selected for its extensive documentation, plentiful examples, and extensive simulation capabilities, making it easier for the developer to get started. The NEST simulation environment was set up in Oracle Virtualbox\footnote{Available online: \protect\url{https://www.virtualbox.org/}, last accessed 06/06/2022}, and Python was used as the programming language, with Jupyter Notebook as the computing platform. Several NEST library packages \cite{NEST} were utilised, including network topology creation, weight setting, and recording device connection to monitor and record spikes generated by the neurons.

The initial step was to determine a method of evaluating the spikes generated by the network. This involved comparing the desired output spike train and the actual spike train generated by the network to measure their deviation and assess the impact of the weight assigned to each synapse on the error.

The main evaluation tool was the \gls*{vp} spike metric, and the \gls*{vr} spike metric was used as a secondary option if the \gls*{vp} distance was not sufficient for comparing the actual and desired spike trains. The Elephant - Electrophysiology Analysis Toolkit\footnote{Available online: \protect\url{https://elephant.readthedocs.io/en/latest/}, last accessed on 10/08/2022} was used to calculate the \gls*{vp} and \gls*{vr} distances.

\gls*{lif} neurons were used to build the network, and various experiments were conducted to understand the behaviour of the neurons based on configurations such as threshold voltage and current. Spike recorders were attached to both the input and output neurons, with the input neuron's spike train serving as the desired output and the output neuron's spike train as the actual output for comparison.

The process of defining the \gls*{snn} topology started with the simplest basic topology and gradually progressed to more complex ones. Neurons were connected by synapses, and their weights were randomly assigned within a predefined range. At first, only excitatory synapses were evaluated, with inhibitory synapses added later. The first network topology used was a 2-layer \gls*{snn} consisting of 2 \gls*{snn} neurons for input and another 2 \gls*{snn} neurons for output, connected by random weights within the range of 10-400, and all synapses were \gls*{stdp} synapses.

\begin{figure}[!t]
\centering
\includegraphics[width=0.9\columnwidth]{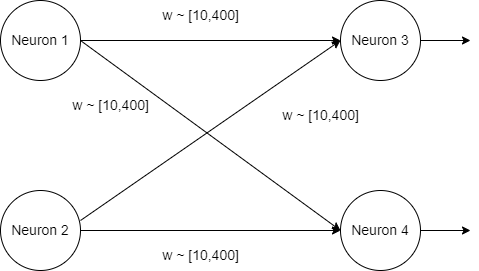}
\caption{An example of a 2-Layer \gls*{snn}.}\label{fig:2layer}
\end{figure}

\begin{figure}[!t]
\centering
\includegraphics[width=\columnwidth]{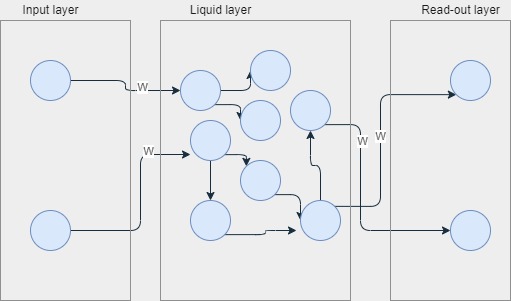}
\caption{A visualisation of the \gls*{lsm} topology.}\label{fig:lsm}
\end{figure}

Fig. \ref{fig:2layer} depicts the network topology of a 2-layer SNN. After conducting experiments and analysing results from the 2-layer network, it was decided to expand the network to a 3-layer topology. The new network consisted of input, middle/hidden, and output layers, each with 5 \gls*{lif} neurons. However, the comparison between the desired and actual spike trains through the use of \gls*{vp} and \gls*{vr} distances did not show a significant difference from the 2-layer network, leading to the decision to increase the number of neurons in each layer to 100. This allowed for further observation of the network's behaviour when different weight ranges were assigned. Building upon the insights gained from the previous experiments, the implementation was further expanded to include the \gls*{lsm} architecture, which can be visualised in Fig. \ref{fig:lsm}.

The defined \gls*{lsm} architecture consists of two neurons in the input layer, eight neurons in the liquid layer, and two neurons in the read-out layer. In the first epoch of the training, the neurons in the liquid layer are connected to each other with random weights within a predefined range. Subsequently, in each epoch, the weights between the input, liquid and the read-out layers respectively are assigned randomly through \gls*{stdp} synapses. The weights and connections within the liquid layer remain unchanged during the training process.

The aim of this approach is to assign weights with a more sensible direction instead of random assignments and to determine a weight range that results in the minimum \gls*{vp} and \gls*{vr} distances. The \gls*{lsm} approach has revealed a range of weights that result in the lowest \gls*{vp} and \gls*{vr} distances. However, the problem with this approach is that the weights are assigned randomly, making it difficult to verify if the weight range is the optimal one for the training process.

To address this issue, weights are initialised using random graphs. The Barabasi-Albert graph is used to generate a random graph of weights within the defined range. The weights are then ordered according to the degree of vertices, with the first $80\%$ of nodes with the highest degree of vertices being assigned as the weights for training in the input layer to the liquid layer and the liquid layer to the read-out layer.

The interconnection of neurons within the liquid layer takes place in the first epoch, with random weights between two randomly selected neurons. A fixed number of inhibitory synapses are added. The input and output neurons are connected directly with a high fixed weight to ensure that the output neuron spikes when the input neuron spikes.

The same weight array is generated using the Erdos Renyi graph and used in the synapses similarly to the Barabasi-Albert graph. The NetworkX python package\footnote{Available online, \protect\url{https://networkx.org/}, last accessed 03/02/2023} is used for implementation and to generate these random graphs.

\section{Results analysis}\label{sec:results}
In this section, the relevant results are discussed. The 2-layer \gls*{snn} topology, consisting of 2 \gls*{lif} neurons per layer, was tested with various weight ranges, and it was found that a significant difference in \gls*{vp} and \gls*{vr} distance was observed throughout the weight range of 10 and 400.

\begin{figure}[!t]
\centering
\includegraphics[width=\columnwidth]{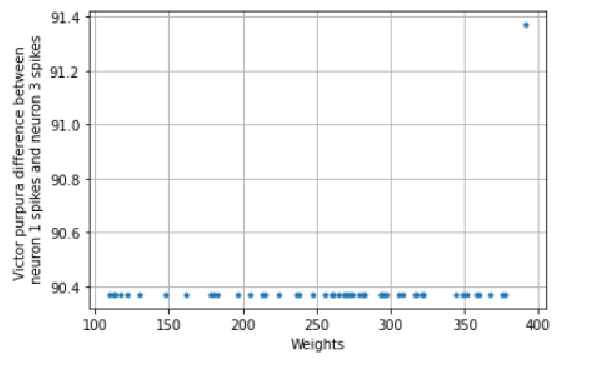}
\caption{2-Layer \gls*{snn} - \gls*{vp} weights difference between input neuron and output neuron spikes. The difference drastically increases when the weight hits 400.}\label{fig:2vp}
\end{figure}

Fig. \ref{fig:2vp} shows the \gls*{vp} difference between input spikes and output spikes. It is clearly visible that the difference drastically increases when the weight hits 400. Weights below 400 result in a minimum \gls*{vp} difference. 

\begin{figure}[!t]
\centering
\includegraphics[width=\columnwidth]{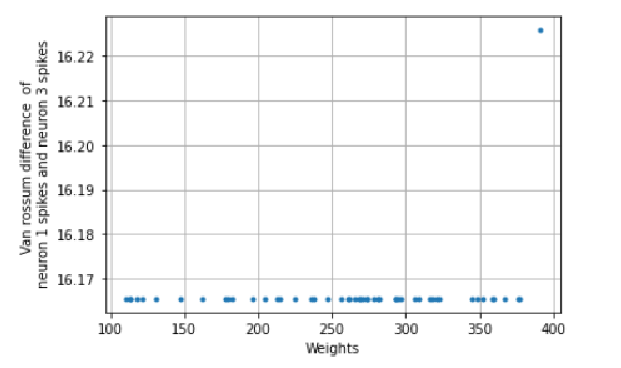}
\caption{2-Layer \gls*{snn} - \gls*{vr} weights difference between input neuron and output neuron spikes. The difference increases when the weight reaches 400.}\label{fig:2vr}
\end{figure}

Fig. \ref{fig:2vr} depicts the \gls*{vr} difference of input and output spikes also shows the same behaviour as Victor Purpura difference. The difference starts to increase when the weight reaches 400. 

However, there was not a noticeable decrease in the distance when the weight was assigned within a certain range. Therefore, the results of the 3-layer \gls*{snn} with a small number of neurons did not show any clear evidence of a weight range that could result in the lowest \gls*{vp} and \gls*{vr} distances. To further explore this issue, a 3-layer \gls*{snn} was implemented with a larger population of neurons. This time, the experiment showed a noticeable decrease in the VP and VR distances when the weights were assigned within a specific range. This result highlights the importance of using a large number of neurons when implementing a 3-layer \gls*{snn} topology.

The experiment also revealed that the weight range can play a crucial role in the performance of \glspl*{snn}. It is essential to find the optimal weight range that can provide the lowest \gls*{vp} and \gls*{vr} distances to enhance the accuracy and efficiency of the \glspl*{snn}.

\begin{figure}[!t]
\centering
\includegraphics[width=\columnwidth]{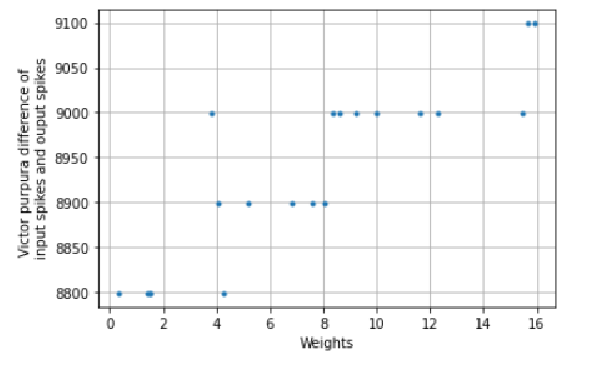}
\caption{3-Layer \gls*{snn} - \gls*{vp} weights difference between input neuron and output neuron spikes. Weights between 0 and 6 have the smallest variation.}\label{fig:3vp}
\end{figure}

\begin{figure}[!t]
\centering
\includegraphics[width=\columnwidth]{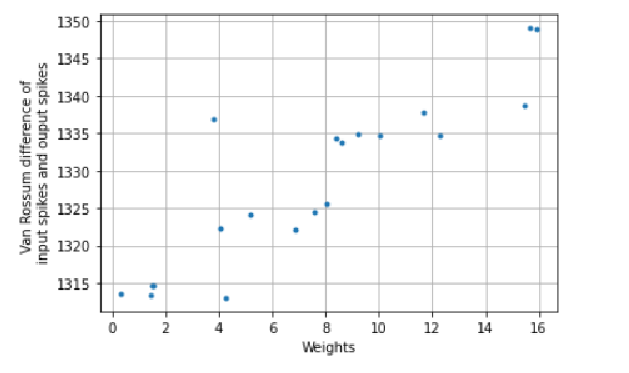}
\caption{3-Layer \gls*{snn} - \gls*{vr} weights difference between input neuron and output neuron spikes. Weights below $6$ are the best range, which results in the lowest difference. }\label{fig:3vr}
\end{figure}

Fig. \ref{fig:3vp} and Fig. \ref{fig:3vr} show the behaviour of input and output spike over the weights, the increase of weight shows a significant \gls*{vp} and \gls*{vr} differences increase. Weights below $6$ are the best range, which results in the lowest difference. 

The results of the above experiments on various network topologies indicated that the discrepancy between the actual and desired output increased with the increase in weight value. This trend was also observed in \gls*{lsm}. Through weight assignment within the range of 1-50, it was observed that the lowest difference was achieved within the weight range of 10-20. In order to improve the results further, the experiment was extended to use a random graph weight initialisation method, the Barabasi-Albert method, instead of purely random weights. The results of this implementation reveals the impact of weight initialisation on the accuracy of the system.

\begin{figure}[!t]
\centering
\includegraphics[width=\columnwidth]{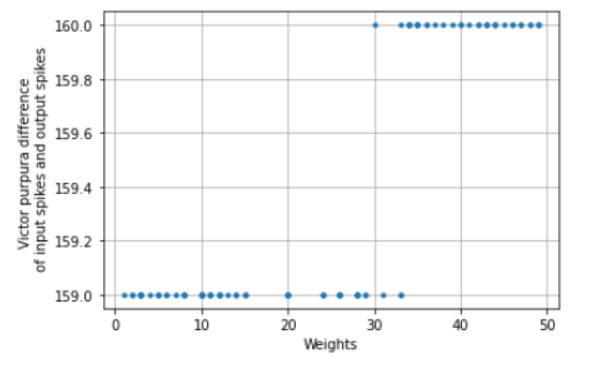}
\caption{\gls*{lsm} - \gls*{vp} weights difference.}\label{fig:lsm-vp}
\end{figure}

\begin{figure}[!t]
\centering
\includegraphics[width=\columnwidth]{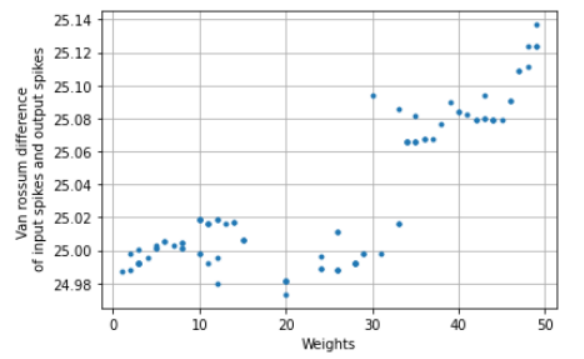}
\caption{\gls*{lsm} - \gls*{vr} weights difference.}\label{fig:lsm-vr}
\end{figure}

From the results depicted in Fig. \ref{fig:lsm-vp} and Fig. \ref{fig:lsm-vr}, it was observed that the Victor Purpura difference of input spikes and output spikes over weight values displayed a more focused distribution compared to the Van Rossum difference graph. The latter demonstrated that the lowest difference was obtained within the weight range of 10-20. Both graphs clearly indicate that the difference in spikes increases as the weight values increase. The same experiment was then repeated using Erdos Renyi as the random graph instead of the Barabasi-Albert graph.

In the final experiment, \gls{lsm} was tested with three main weight initialization approaches for various weight ranges, including 1-10, 1-20, 1-50, and 1-100, among others. The behaviours of the \gls*{vp} and \gls*{vr} differences were closely monitored for each weight range, which was initialised using purely random initialisation, the Barabasi-Albert graph, and the Erdos Renyi graph. Table \ref{tab:table} showcases the lowest difference achieved for each weight range using the three main weight initialisation methods.

The highlighted rows are the minimum \gls*{vr} difference obtained in each weight range. As per the above experiment, it is clearly visible that the majority of the least difference was obtained through Barabasi-Albert random graphs and the best weight range for all the experiment results is below 20.

\begin{table*}[h]
    \centering
        \caption{Experimental run of obtaining least \gls*{vp} and \gls*{vr} difference using pure random, Barabasi-Albert and Erdos Renyi graphs.}\label{tab:table}
    \begin{tabular}{c}
         \includegraphics[width=1.0\textwidth]{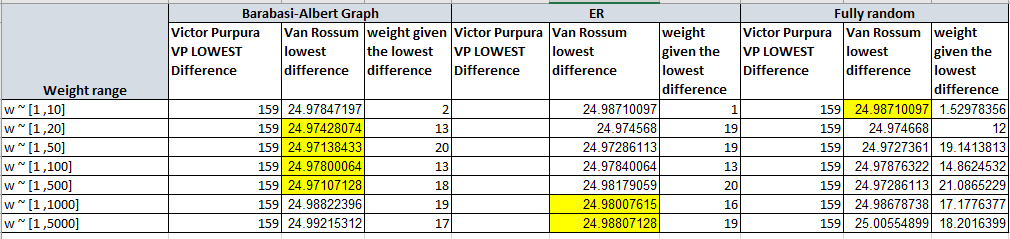} \\
    \end{tabular}
\end{table*}

\section{Conclusion and Future Work}\label{sec:conclusions}
Based on the results of the research, it can be concluded that the difference between the input spike train and output spike train, as measured by both \gls*{vp} and \gls*{vr} distances, increases as the weights of the synapses increase. The Victor Purpura distance has a clearer distinction when the neuron population is large, and for such populations, the relationship between weights and the \gls*{vp} or\gls*{vr} distances follows an exponential pattern.

In the \gls*{lsm} experiments, the \gls*{vp} distance remained constant, while the \gls*{vr}difference was affected by the weight range. This highlights the importance of using multiple spike metrics when observing network behaviour. The relationship between weights and spike metric differences was consistent across all network topologies tested in the research. The identified weight range for LSM topology is the optimal range for assigning excitatory synapses in order to minimise the spike metric difference. As further research, the weight range for inhibitory synapses can be determined in order to result in a similar spike train to the input spike train. The current implementation does not take into account lateral inhibition, so this will be a future area of research to observe the impact of this property on training optimisation. Additionally, the propagation delays of pre-synaptic neurons have not been considered in the current research, so this will also be a future area of investigation to improve the accuracy of the solution.

From a societal perspective, the findings of this research have implications for the broader field of \gls*{ai}. Understanding the relationship between synaptic weights and spike metric differences contributes to the development of more efficient and accurate neural network models. These advancements can positively impact various \gls*{ai} applications, such as image recognition, natural language processing, and autonomous systems, leading to improved performance and reliability. Moreover, the optimization of training processes based on these findings can reduce computational resources and energy consumption, making \gls*{ai} systems more sustainable. In terms of novel exploitation and business models, the research outcomes offer opportunities for developing specialised \gls*{ai} solutions. Companies can leverage the insights gained from this research to create advanced neural network architectures with optimized synaptic weights. Such innovations can be applied in areas like anomaly detection, pattern recognition, and predictive analytics, enabling businesses to make more informed decisions and enhance operational efficiency. Additionally, organizations specializing in \gls*{ai} hardware and software can incorporate these findings into their products and services, providing customers with more efficient and accurate \gls*{ai} tools.

The current experiments have only been performed with small populations of neurons. It is expected that noise will increase in larger populations, so it will be important to incorporate a mechanism for reducing noise in future work. Furthermore, the current approach can be extended to consider other important factors that can impact the training optimisation process of \gls*{lsm}, such as lateral inhibition, propagation delays of pre-synaptic neurons, noise reduction mechanisms, and the impact of larger neuron populations. Additionally, the solution of this research can be developed into a package or library that can be used by researchers and developers to optimise the range of weights for different \gls*{snn} topologies. This package should be able to provide the optimal weight range based on inputs such as network topology, number of neurons per layer, number of synapses, and type of synapses.

\bibliographystyle{ieeetr}
\bibliography{references}

\end{document}